\documentclass[a4paper,fleqn]{cas-dc}
\usepackage[ruled,linesnumbered]{algorithm2e}
\usepackage{algpseudocode}
\usepackage{graphicx}
\usepackage{booktabs}
\usepackage{multirow}
\usepackage{caption}
\usepackage{subcaption}
\usepackage{geometry}
\usepackage[section]{placeins}
\usepackage{float}
\usepackage{amsmath}
\newcommand{\gr}{\rowcolor[gray]{.95}}
\usepackage{graphicx}

\def\tsc#1{\csdef{#1}{\textsc{\lowercase{#1}}\xspace}}
\tsc{WGM}
\tsc{QE}
\tsc{EP}
\tsc{PMS}
\tsc{BEC}
\tsc{DE}

\shorttitle{Performant Anatomical Landmark Detection}
\shortauthors{X. Zhou et~al.}
\title [mode = title]{HYATT-Net is Grand: A Hybrid Attention Network for Performant Anatomical Landmark Detection}  

\tnotetext[1]{X. Zhou and Z. Huang contribute equally to this work.}

\author[1, 2]{Xiaoqian Zhou}
\affiliation[1]{organization={School of Biomedical Engineering, Division of Life Sciences and Medicine, University of Science and Technology of China(USTC)},
                city={Suzhou},
                postcode={230026},
                country={China}}
                
\affiliation[2]{organization={Suzhou Institute for Advanced Research, University of Science and Technology of China (USTC)},
                city={Suzhou},
                postcode={215123}, 
                country={China}}

\author[3,4]{Zhen Huang}[style=chinese]
\affiliation[3]{organization={School of Computer Science and Technology, University of Science and Technology of China (USTC)},
                city={Hefei},
                postcode={230026}, 
                country={China}}
                
\affiliation[4]{organization={School of Information Science and Technology, Eastern Institute of Technology (EIT)},
                city={Ningbo},
                postcode={315200}, 
                country={China}}

\author[1, 2]{Heqin Zhu}[style=chinese]

\author[5]{Qingsong Yao}[style=chinese]
\affiliation[5]{organization={Stanford University},
                city={Palo Alto, California},
                postcode={94305},
                country={United States}}
                
\author[1, 2]{S.Kevin Zhou}[style=chinese]
\orcidauthor{0000-0002-6881-4444}{SK.Zhou \ead{skevinzhou@ustc.edu.cn}}

\date{}  

\begin{document}

\begin{abstract}
Anatomical landmark detection (ALD) from a medical image is crucial for a wide array of clinical applications. While existing methods achieve quite some success in ALD, they often struggle to balance global context with computational efficiency, particularly with high-resolution images, thereby leading to the rise of a natural question: where is the performance limit of ALD? In this paper, we aim to forge performant ALD by proposing a {\bf HY}brid {\bf ATT}ention {\bf Net}work (HYATT-Net) with the following designs: (i) A novel hybrid architecture that integrates CNNs and Transformers. Its core is the BiFormer module, utilizing Bi-Level Routing Attention for efficient attention to relevant image regions. This, combined with Attention Residual Module(ARM), enables precise local feature refinement guided by the global context. (ii) A Feature Fusion Correction Module that aggregates multi-scale features and thus mitigates a resolution loss. Deep supervision with a mean-square error loss on multi-resolution heatmaps optimizes the model. Experiments on five diverse datasets demonstrate state-of-the-art performance, surpassing existing methods in accuracy, robustness, and efficiency. The HYATT-Net provides a promising solution for accurate and efficient ALD in complex medical images. Our codes and data are already released at: \url{https://github.com/ECNUACRush/HYATT-Net}.
\end{abstract}


\maketitle

\section{Introduction}
Anatomical landmark detection (ALD) is a core task in medical image analysis, widely used in clinical diagnosis, surgical planning, and treatment evaluation~\cite{landmarkdetection1,landmarkdetection2}. Accurate landmark localization provides reliable anatomical references for clinicians, enabling automated diagnosis and personalized treatment~\cite{treatment}. It also supports foundational tasks such as image registration~\cite{registration}, segmentation~\cite{forests, shao2023diffuseexpand}, and 3D reconstruction~\cite{3Dreconstruction}, improving the accuracy of these tasks and advancing medical imaging technology. However, complex anatomical structures, individual variability, low signal-to-noise ratio and resolution differences in medical images make automated ALD challenging.

Early ALD methods rely on handcrafted image features~\cite{handcrafted1,handcrafted2,handcrafted3}. While these methods achieve some success in specific applications, they depend heavily on data quality and feature design. As a result, they struggle to handle complex anatomical structures and adapt to varying imaging conditions, leading to limited robustness. With the rapid advancement of deep learning, Convolutional Neural Networks (CNNs) have transformed image processing tasks, particularly in ALD and medical image segmentation. Models such as U-Net~\cite{Unet} and its variants~\cite{CNN1, CNN2, oNeil, SCN, FARNet} are now widely applied for landmark detection. These models employ encoder-decoder architectures to effectively integrate low-level features with high-level semantic information.

Transformer architectures, with their self-attention mechanisms, are particularly effective at modeling global dependencies and capturing long-range spatial information. In the field of medical imaging, Transformers~\cite{vaswani2017attention, VIT} have been increasingly integrated into image analysis tasks, often in combination with convolution-based U-Net models such as UTNet~\cite{utnet}, TransUNet~\cite{transunet}, and nnFormer~\cite{nnformer}. Since accurate ALD relies heavily on global context, incorporating Transformers to capture this information has been shown to enhance detection accuracy. Notable examples include SpineHRformer~\cite{SpineHRformer} and DATR~\cite{DATR}. 

To leverage the complementary strengths of Convolutional Neural Networks (CNNs) and Transformers, hybrid architectures such as HTC~\cite{HTC} and CephalFormer~\cite{cephalformer} have emerged. These models combine CNNs for local feature extraction and Transformers for modeling global dependencies, achieving significant success in ALD. However, the computational complexity of standard self-attention mechanisms scales quadratically with the number of tokens, which limits the computational efficiency and overall performance of the model to some extent. Some methods, such as Gated Axial UNet (MedT)~\cite{Gated}, have introduced sparsity to alleviate the computational burden, but they rely on manually designed, fixed sparse patterns. These fixed patterns limit the model's adaptability when processing different queries, as they cannot dynamically adjust the sparse pattern according to the specific context. Medical images are typically characterized by a high complexity, a low signal-to-noise ratio, and a significant domain gap between datasets from different anatomical regions and acquisition protocols. These characteristics demand models that are able to flexibly select and adjust computation regions to address these challenges. Therefore, developing a sparse attention mechanism that dynamically adjusts with queries tailored for medical image landmark detection tasks remains one of the key research directions to improve detection accuracy.

To address the above challenges and forge performant ALD with accuracy and robustness, we propose a novel {\bf HY}brid {\bf ATT}ention {\bf Net}work (HYATT-Net), which incorporates a dynamic sparse attention into a hybrid Transformer-CNN architecture, specifically designed for ALD. Our approach leverages an \textbf{Attention Residual Module (ARM)} and a \textbf{BiFormer module}, marking the first combination in ALD to achieve state-of-the-art (SOTA) results. Specifically, the ARM enhances feature selection through channel and spatial attention mechanisms, while the BiFormer module leverages bi-level routing attention to effectively model long-range spatial dependencies with optimized computational and memory efficiency. Additionally, we design a Feature Fusion Correction Module (FFCM) to integrate global and local features, further improving detection accuracy and adaptability to diverse anatomical structures and imaging conditions. Deep supervision is also employed to refine predictions at multiple levels, enhancing robustness and performance.

We conduct extensive experiments across five public datasets, encompassing various anatomical regions, imaging methods, and resolutions, consistently achieving state-of-the-art (SOTA) results. For example, on the ISBI2015 dataset, our method achieves a mean residual error (MRE) of 1.13mm, improving by 5\% compared to HTC~\cite{HTC} (1.19mm) and 11\% compared to FARNet~\cite{FARNet} (1.27mm). Additionally, the Success Detection Rate (SDR) at 2mm reaches 84.78\%, surpassing HTC by 1.54\% and FARNet by 2.27\%. Similar improvements are observed across all datasets, demonstrating the robustness, adaptability, and clear superiority of our method in high-resolution medical image analysis and landmark detection tasks.

In summary, the main contributions of this paper are as follows:
\begin{itemize}
    \item We propose a novel architecture that first integrates the newly designed BiFormer module with the Attention Residual Module (ARM) to enhance feature selectivity and capture global dependencies, thereby improving the accuracy and robustness of Anatomical landmark detection (ALD).
    \item We incorporate deep supervision during training, along with the Feature Fusion Correction Module (FFCM), to effectively integrate global and local information, further enhancing adaptability to diverse anatomical regions and imaging modalities.  
    \item Comprehensive experiments on multiple public medical datasets, including ISBI2015, Hand, and ISBI2023, demonstrate that our method achieves state-of-the-art (SOTA) performance, showcasing its strong adaptability and robustness across diverse datasets.
\end{itemize}




\section{Related Work}
\subsection{Traditional ALD Approaches}
Landmark detection methods can be broadly classified into two categories: heatmap prediction and direct regression. Heatmap-based methods estimate the likelihood of landmark positions by generating probability maps, while regression-based methods directly predict the coordinates of landmarks~\cite{regression}. Given the significance of contextual information in medical imaging, heatmap-based approaches are generally more suitable for landmark detection in medical applications than regression-based methods.

Traditional feature extraction methods include image filters like SIFT~\cite{SIFT} to extract invariant features. Rule-based methods~\cite{ruled1,ruled2} detect edges and contours using image processing techniques and prior landmark knowledge. Liu et al.~\cite{liu} propose a submodular optimization framework that leverages spatial relationships among landmarks to enhance detection accuracy. However, as image complexity increases, these methods become challenging to maintain. Later, some studies employ template matching~\cite{template1,template2} and introduce Active Shape Models (ASM) and Active Appearance Models (AAM)~\cite{ASM, AAM} to improve the effectiveness of landmark detection.


Although these methods achieve some success in ALD, they gradually fall short due to the high complexity and precision requirements of medical imaging. With the rapid development of deep learning, researchers begin exploring new approaches to overcome the limitations of traditional methods, significantly improving the performance and adaptability of landmark detection.

\subsection{CNN or Transformer Based ALD Methods}
With the rapid advancements in deep learning, Convolutional Neural Networks (CNNs) have been widely adopted for ALD. For instance, O'Neil et al.~\cite{oNeil} propose a two-stage method based on Fully Convolutional Networks (FCN). Payer et al.~\cite{SCN} introduce Spatial Configuration Network (SCN), which combines local appearance features with spatial configuration information. Zhu et al.~\cite{GU2Net} propose GU2Net, a general model for multi-domain landmark detection, which improves robustness through cross-domain learning. Additionally, Ao et al.~\cite{FARNet} develop FARNet, which incorporates multi-scale feature aggregation and refinement modules, further improving detection precision.  

However, traditional CNN architectures often struggle to capture long-range dependencies, limiting their effectiveness in handling complex anatomical structures for landmark detection tasks. To address these limitations, Transformer architectures have been increasingly adopted for landmark detection. Unlike traditional CNN-based methods that primarily focus on local spatial relationships, Transformers leverage self-attention mechanisms to model global dependencies and capture long-range information. For example, SpineHRformer~\cite{SpineHRformer} employs Transformer modules for detecting spinal anatomical landmarks, achieving significant success in landmark detection. Zhu et al. introduce Domain Adaptive Transformer (DATR)~\cite{DATR}, which utilizes Transformers and domain-adaptive blocks to capture global dependencies and domain-specific features, further enhancing landmark detection performance. By combining the strengths of Transformers and Encoder-Decoder structure, global context and spatial dependencies are effectively captured, enabling accurate detection of anatomical landmarks in chest X-ray images~\cite{chest}. These results highlight the potential of Transformers in medical image landmark detection and provide strong support for their further application in medical imaging analysis.
\subsection{CNN-Transformer Hybrid Models}
To fully exploit the strengths of both CNNs and Transformers, various hybrid architectures have emerged in recent years. These models combine the local feature extraction capabilities of CNNs with the global modeling power of Transformers, driving significant progress in landmark detection tasks. For example, CephalFormer~\cite{cephalformer} integrates interleaved convolutional and Transformer blocks, enabling both coarse and fine-grained landmark detection in two stages, and excels in both 2D and 3D landmark detection tasks. HTC~\cite{HTC} combines CNNs and Transformers, utilizing convolutional encoders to extract local features and Transformer encoders to capture global context and long-range dependencies, achieving superior performance in landmark detection. 

However, due to the complexity and low signal-to-noise ratio of medical images, high-performance models are necessary. The computational complexity of standard Transformer models becomes a bottleneck, especially when processing high-resolution medical images. Some studies, such as Swin-Unet~\cite{SwinUnet} and Gated Axial UNet (MedT)~\cite{Gated}, have incorporated sparsity into their models to address this challenge. Swin-CE~\cite{SwinCE} integrates Swin Transformer~\cite{liu2021swin} encoders with convolutional encoders for cephalometric landmark detection, but these methods often rely on manually designed sparse patterns, which lack flexibility in adapting to dynamic queries.

Recent advancements in vision Transformers, particularly those utilizing dynamic sparsity, provide promising solutions. For instance, dynamic token sparsity mechanisms prune a large number of non-informative tokens, thereby accelerating the model while maintaining accuracy. In~\cite{biformer}, dynamic token sparsity is applied to reduce computational load by removing non-informative tokens without sacrificing precision. The application of dynamic and query-aware sparse attention mechanisms for landmark detection in medical images presents a significant research opportunity. In this study, we propose employing a bi-level routing attention as a foundational dynamic sparse block, constructing a U-shaped hybrid CNN-Transformer encoder-decoder architecture for efficient landmark detection in medical images.

\section{Proposed Methods}
In this section, we first define the landmark detection problem in Section 3.1. Section 3.2 introduces the BiFormer module and its design. Section 3.3 covers the overall network architecture, incorporating the BiFormer module into a hybrid CNN-Transformer framework. Lastly, Section 3.4 discusses the loss function with deep supervision learning strategy. 

\begin{figure*}[htbp]
\centering
\begin{minipage}{\textwidth}
  \centering
  \includegraphics[width=0.8\linewidth,trim={0cm 0cm 0cm 0cm},clip]{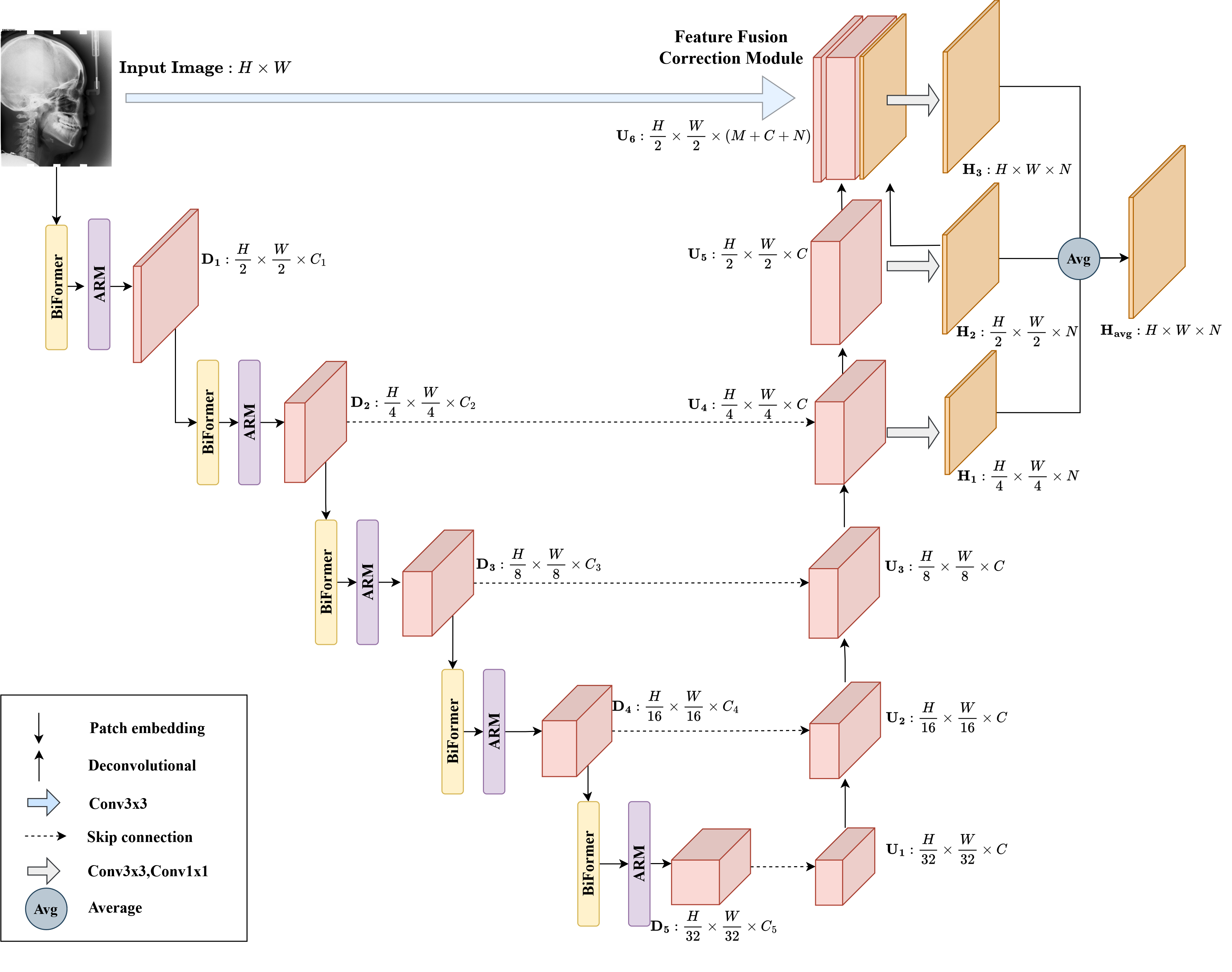}
  \caption{The overview of proposed Hybrid Attention Network(HYATT-Net). BiFormer is a module based on bilevel routing attention, and ARM stands for a Attention Residual Block. Further details will be discussed later. $N$ denote the number of landmarks.}
  \label{fig: network}
\end{minipage}
\end{figure*}

\subsection{Problem Definition}
The task of ALD aims to accurately predict the locations of multiple anatomical landmarks from input medical images. Formally, given an input image \( I \in \mathbb{R}^{H \times W \times C} \), the objective is to predict a set of landmarks \( L = \{l_1, l_2, \dots, l_N\} \), where each \( l_i = (x_i, y_i) \) represents the position of a landmark in the image. \( N \) represents the number of landmarks in the image, which is also the number of output heatmap channels. The corresponding heatmap \( H_i(x_i, y_i) \) for each landmark \( l_i \) is generated using a Gaussian function:
\begin{equation}
\begin{aligned}
    H_{i} = \frac{1}{\sqrt{2\pi} \sigma} \exp \left( -\frac{(x - x_{i})^2 + (y - y_{i})^2}{2\sigma^2} \right),
\end{aligned}
\end{equation}
where \( C \) represents the number of image channels, with \( C=1 \) for X-ray images, \( H \) and \( W \) represent the image height and width, respectively, and \( \sigma \) represents the standard deviation of the Gaussian function used to generate the heatmap.

\subsection{BiFormer Module Design}
The BiFormer module, based on a Bi-Level Routing Attention (BRA) mechanism~\cite{biformer}, is a key component of the proposed architecture. BRA is a dynamic, query-aware sparse attention mechanism designed to prune the least relevant key-value patchs at a coarse level, retaining only the most relevant parts for fine-grained token-level attention. Token-to-token attention is then computed within these selected patches. Compared to static sparse attention mechanisms~\cite{SwinUnet,DAT}, BRA is more flexible in modeling long-range dependencies, making it particularly effective for dense prediction tasks that require capturing global features.

\begin{figure}[htbp]
\centering
\includegraphics[width=\columnwidth, trim={0cm 0cm 0cm 0cm}, clip]{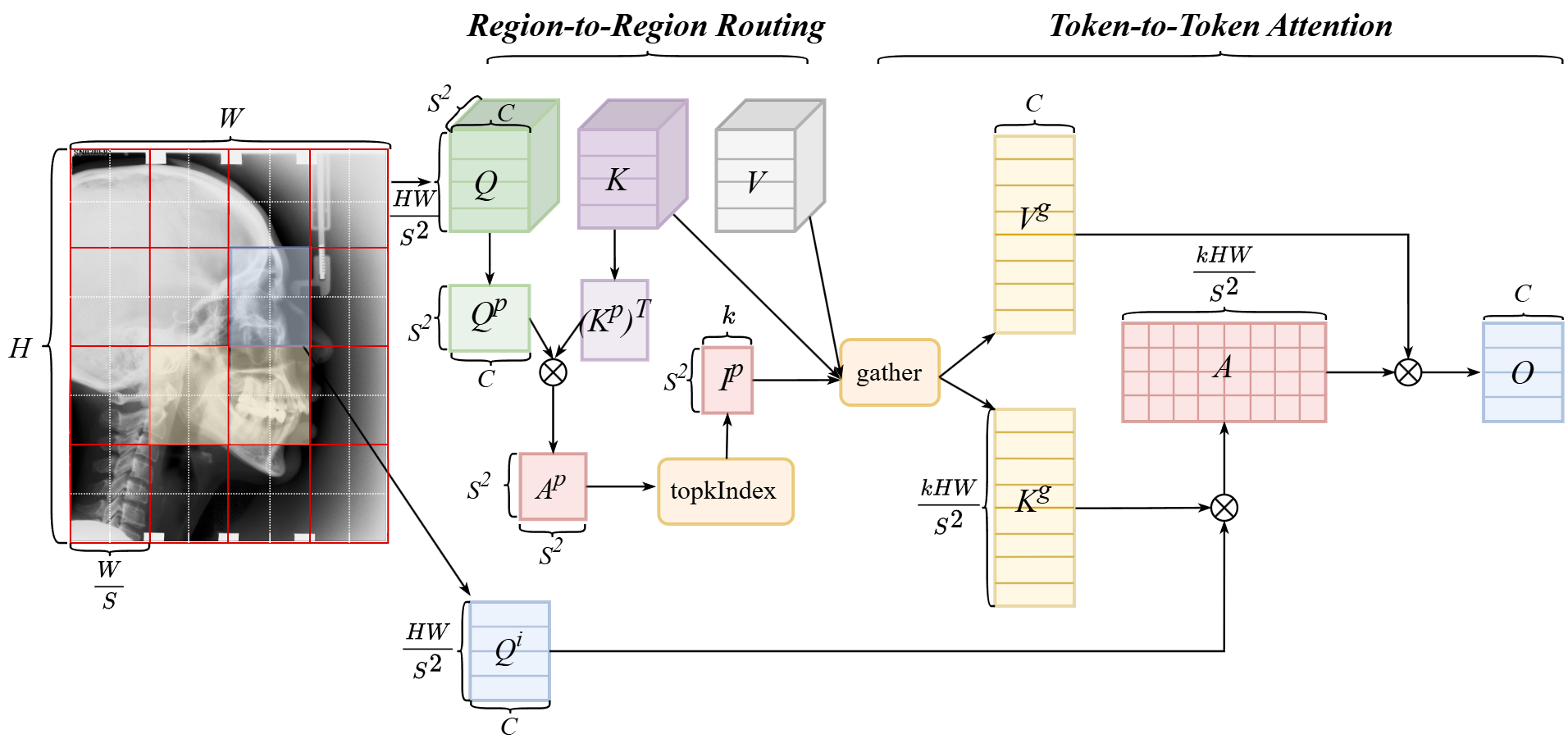}
\caption{Illustration of region-to-region routing and token-to-token attention. Our approach leverages sparsity by gathering key-value pairs from the top-\( k \) related windows, bypassing irrelevant computations and focusing on GPU-friendly dense matrix multiplications for improved efficiency.}
\label{fig: biformer}
\end{figure}

Given a 2D X-ray medical image \( X \in \mathbb{R}^{H \times W \times C} \), where \( C = 1 \), the BiFormer block partitions the image into non-overlapping patches of size \( S \times S \), with each patch having a feature size of \( \frac{HW}{S^2} \). These patches are then linearly projected to compute the query (\( Q \)), key (\( K \)), and value (\( V \)) matrices for all patches. The resulting \( Q \), \( K \), and \( V \) are represented as \( Q, K, V \in \mathbb{R}^{S^2 \times \frac{HW}{S^2} \times C} \). Moreover, as illustrated in Fig.~\ref{fig: biformer}, the average values of \( Q \) and \( K \) for each patch are computed to obtain \( Q^p \) and \( K^p \), where \( Q^p, K^p \in \mathbb{R}^{S^2 \times C} \). A matrix multiplication between \( Q^p \) and the transposed \( K^p \) is then performed, resulting in the patch-to-patch adjacency matrix \( A^p \), where \( A^p \in \mathbb{R}^{S^2 \times S^2} \). This matrix \( A^p \) encodes the degree of association between each pair of patches. Next, a row-wise top-\( k \) operator, denoted as \( topkIndex() \), is applied to the adjacency matrix to identify the top-\( k \) most relevant patches, producing an index matrix \( I^p \), where \( I^p \in \mathbb{N}^{S^2 \times k} \). This index matrix \( I^p \) captures the most relevant patches for each patch in the image.


Furthermore, as shown in Fig.~\ref{fig: biformer}, to compute the token-to-token attention for a patch \( i \), initially, collect the top-\( k \) most relevant patch index from matrix \( I^p \) of patch \( i \); subsequently, apply the gather algorithm to extract the corresponding \( K^g \) and \( V^g \) values of the relevant patches, where \( K^g, V^g \in \mathbb{R}^{S^2 \times k \frac{HW}{S^2} \times C} \). Finally, calculate the attention between patch \( i \) and the top-\( k \) most relevant patches, incorporating the Local Context Enhancement (LCE), to obtain the final output \( O \) for this patch. The above process can be formulated as follows:
\begin{equation}
\begin{aligned}
\label{eq:BRA3}
K^g = \text{gather}(K, I^p), \quad V^g = \text{gather}(V, I^p),
\end{aligned}
\end{equation}
\vspace{-6mm} 
\begin{equation}
\label{eq:BRA4}
O = \text{softmax} \left( \frac{Q (K^g)^T}{\sqrt{C}} \right) V^g + \text{LCE}(V),
\end{equation}
where $\text{LCE}(V)$ is parametrized with a depth-wise convolution, with kernel size set to 5.

\begin{figure}[htbp]
\centering
\includegraphics[width=\columnwidth, trim={0cm 0cm 0cm 0cm}, clip]{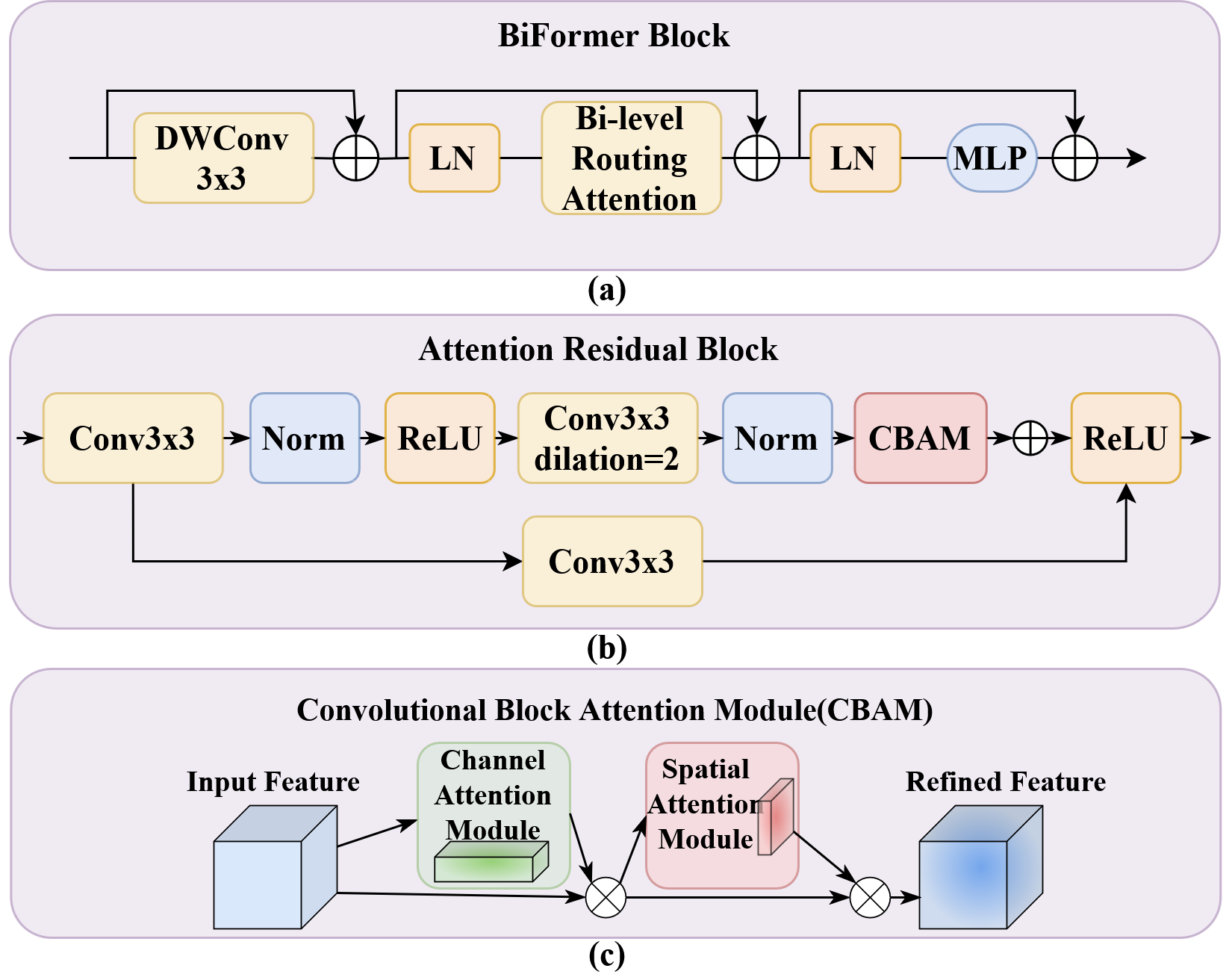}
\sloppy  
\caption{(a) Architecture of the BiFormer Block. (b) Architecture of the proposed Attention Residual Block. (c) Overview of the Convolutional Block Attention Module (CBAM).}
\label{fig: Block}
\end{figure}

As demonstrated in Fig.~\ref{fig: Block}(a), BiFormer can effectively capture global features through the BRA mechanism, particularly the long-range dependencies between landmarks and overall geometric relationships. The input feature is firstly processed through a $3\times 3$ depth-wise convolution (DWConv) to encode relative positional relationships implicitly. The BRA module then learns cross-position relationships from the input, followed by an MLP layer that further processes the feature representation at each position.

Compared to the traditional fully connected global attention mechanism, BiFormer not only reduces redundant computations but also preserves the effectiveness and accuracy of global features in complex structures and multi-scale scenarios. Moreover, adaptive sparse attention enables the model to dynamically select features to focus on based on the input content, which is essential for accurate landmark detection in complex anatomical structures.

\subsection{Overal Network Architecture}
As demonstrated in Fig.~\ref{fig: network}, the overall HYATT-Net employs a U-shaped architecture, with the main backbone consisting of hybrid Transformer-CNN modules, upsampling layers, and residual connections. The hybrid Transformer-CNN module is composed of BiFormer and ARM connected in series. The global features extracted by BiFormer are fed into ARM, which further refines local features, achieving an effective integration of global and local features. Specifically, the global information provided by BiFormer helps the model understand the geometric relationships between landmarks, while ARM refines local features based on these global features, making landmark detection more precise.

Compared to the original CNN, ARM is integrated with an attention module Convolutional Block Attention Module(CBAM)~\cite{woo2018cbam} to select representative features. As demonstrated in Fig.~\ref{fig: Block}(c), the CBAM module combines channel attention and spatial attention, allowing the network to focus more effectively on task-relevant features, enhancing the effectiveness of feature representation and the overall performance of the network. The role of CBAM can be formulated as:
\begin{align*}
F_{\text{out}} = \text{SA}(\text{CA}(F_{\text{in}}) \odot F_{\text{in}}) \odot (\text{CA}(F_{\text{in}}) \odot F_{\text{in}}),
\end{align*}
where CA represents channel attention, SA represents spatial attention, and \( \odot \) denotes element-wise multiplication.

As for input features, ARM first processes them using a dilated convolution with dilation set to 1, followed by normalization and activation function processing. Then, the processed features are passed through another dilated convolution with dilation set to 2. After normalization, the features are passed through the CBAM, and finally, the output is obtained through a residual connection. The above process illustrated in Fig.~\ref{fig: Block}(b) and can be represented as:
\begin{align*}
F_{\text{out}} =&\ \text{RELU} ( \text{Conv} ( F_{\text{in}} ) + \text{CBAM} (  \\ 
&\ \text{Norm} ( \text{Conv} ( \text{RELU} ( \text{Norm} ( \text{Conv} ( F_{\text{in}} )  )  )  )  )  )  ) ,
\end{align*}


BiFormer provides rich contextual information to ARM, reducing redundant features and information loss, thereby accelerating the training process and improving convergence speed. Therefore, combining BiFormer with ARM not only enhances the model's performance in landmark detection but also improves training efficiency and model robustness, making it suitable for complex medical images and other scenarios that require precise modeling.

As Fig.~\ref{fig: network} shows, the downsampling part of the model consists of five layers of hybrid Transformer-CNN modules. For an input image, Patch embedding is first performed, followed by the BiFormer module for capturing global features and the ARM module for extracting local features. Multi-scale features \(\{D_1, D_2, D_3, D_4, D_5\}\) are generated to enhance the model's feature representation capability. These features are subsequently mapped to a feature space with 256 channels.

During the upsampling process, we upsample the features using a de-convolution operation and retain the original detail information through residual connections to maintain landmark precision, resulting in \(\{U_1, U_2, U_3, U_4, U_5\}\). The features are passed through $3\times 3$ and $1\times 1$ convolution operations to obtain the predicted heatmap. Since the first layer of the local network is not used during the upsampling phase, the final generated heatmap is in the shape of the downsampled image, forming the loss of information at the original image resolution.

To address the above issue, we introduce a Feature Fusion Correction Module (FFCM) that is responsible for extracting the overall contextual information of the input image. This module uses global pooling layers and fully connected networks to generate image features, aiding the model in better capturing the relative relationships between anatomical structures. The extracted features are concatenated with the upsampled feature map \( U_5 \) and the heatmap \( H_2 \) to generate the feature map \( U_6 \), which is then passed through \( 3 \times 3 \) and \( 1 \times 1 \) convolution operations to obtain the predicted heatmap \( H_3 \). Through the FFCM, the model can not only effectively restore the resolution loss caused by downsampling but also better integrate global and local information, ensuring more stable and accurate performance in complex anatomical structures. This enhancement makes the model more adaptable and generalizable to various medical imaging applications, particularly demonstrating significant performance improvements in detail-sensitive landmark detection tasks.

\subsection{Loss Function with Deep Supervision Strategy}
Detecting and segmenting abnormal regions, such as tumours, can benefit from heatmaps at different resolutions, which assist in identifying the size, shape, and position of the tumour. Therefore, we introduce a deep supervision learning strategy during model training to generate multi-resolution heatmaps, aiming to improve landmark detection accuracy. Specifically, we progressively generate heatmaps at different scales, from low to high resolution, and apply heatmaps with sigma values that decrease from large to small, facilitating coarse-to-fine landmark detection. This multi-scale supervision strategy enhances the model's ability to leverage information across different scales, improving its robustness and generalization in complex anatomical structures.

Additionally, we use Mean Squared Error (MSE) loss to measure the difference between the predicted heatmap and the ground truth. By penalizing the model at each scale of the multi-resolution heatmaps, the model maintains high prediction accuracy across different scales. The loss function is defined as follows:
\begin{equation}
\hspace{1cm} \mathcal L_{\text{all}} = \mathcal L_{H_1} + \alpha_1 \mathcal L_{H_2} + \alpha_2 \mathcal L_{H_3},
\end{equation}

In the equation, \( \mathcal{L}_{\text{all}} \) represents the total loss function for model training. The loss functions \( \mathcal{L}_{H_1} \), \( \mathcal{L}_{H_2} \), and \( \mathcal{L}_{H_3} \) correspond to the landmarks in the heatmaps \( H_1 \), \( H_2 \), and \( H_3 \), which are the outputs of the deep supervision layers, as shown in Fig.~\ref{fig: network}. \( \alpha_1=3 \) and \( \alpha_2=3 \) are the weights of the losses, which can be adjusted according to experimental needs. Finally, the output of the deep supervision layers is averaged to obtain the final prediction of the model, thereby further improving the accuracy and stability of landmark localization.

\section{Experiment}
\subsection{Datasets and Metrics}
\subsubsection{Datasets}
In this study, we evaluate our method using five different datasets, including three head X-ray datasets (two public datasets and one private dataset), one public hand X-ray dataset, and one public pelvic X-ray dataset. Our method demonstrates SOTA results across these various datasets under most metrics, indicating strong robustness and broad applicability. Details of each dataset are as follows:

\textbf{ISBI 2015:}
This public cephalometric dataset~\cite{ISBI2015} is from the ISBI 2015 Grand Challenge, consisting of 150 training images and two test sets, totaling 250 images. In accordance with standard practices, the two test sets are combined for model evaluation. Each X-ray image has a resolution of 1935x2400 with 0.1mm x 0.1mm pixel spacing. The images are uniformly resized to 1024x1216 for network training. Each image is annotated with 19 landmarks by two experts, and the average of their annotations is used as the ground truth.

\textbf{ISBI 2023:}
This cephalometric dataset~\cite{ISBI2023}, part of the IEEE ISBI 2023 Challenge, contains 700 X-ray images captured by seven different imaging devices. Each image is annotated with 29 landmarks by two experts, and the average of their annotations is used as the ground truth. Due to variations in image size and resolution across the devices, all images are resized to 1024x1216. The dataset is split such that 75\% of the images are used for training and 25\% for testing.

\textbf{CephAdoAdu:}
This private cephalometric dataset~\cite{CELDA} comprises 700 head X-ray images, evenly divided into 350 adult images and 350 adolescent images, with notable visual differences between the two groups. Dental experts manually annotate 10 landmarks per image. The dataset is split into 200 training images and 150 test images for both adults and adolescents, totaling 400 training images and 300 test images. The image spacing is 0.1mm, and due to varying image sizes, all images are resized to 1024x1024 for consistency.

\textbf{Hand X-Rays\footnote{https://ipilab.usc.edu/research/baaweb/}:}
This public dataset contains 895 X-ray images, with 37 annotated hand landmarks per image. The images are resized to 1024x1216 and split into 75\% for training and 25\% for testing. Due to the lack of spacing information, we assume a distance of 50mm between the landmarks at both ends of the wrist, consistent with previous work~\cite{SCN} to estimate the actual landmark distances.

\textbf{Pelvic X-Rays\footnote{https://www.kaggle.com/datasets/tommyngx/cgmh-pelvisseg}:}
This dataset is sourced from the CGMH-PelvisSeg public dataset, which includes 400 high-resolution pelvic X-ray images, along with an additional 150 images from another public dataset~\cite{Pelvic}. Based on the PELE setup, 132 images are selected for experiments, with 107 used for training and 25 for testing. To ensure a fair comparison with the original work and other methods, all images are resized to 512x512 following PELE~\cite{PELE}’s specifications. Additionally, due to the unknown pixel spacing, model performance is evaluated by calculating pixel distances between landmarks in the images.

\subsubsection{Evaluation Metrics}
To assess the model's performance, we use two commonly employed metrics: MRE and SDR.

\textbf{Mean Radial Error (MRE):}
This metric calculates the average Euclidean distance between the true and predicted landmarks, where a lower MRE signifies better performance.

\begin{equation}
\hspace{0.5cm} R = \sqrt{(x^{\text{pred}} - x^{\text{gt}})^2 + (y^{\text{pred}} - y^{\text{gt}})^2},
\end{equation}
\vspace{-6mm} 
\begin{equation}
\hspace{2cm} \text{MRE} = \frac{\sum_{i=1}^{N} R_i}{N},
\end{equation}
where \(R\) represents the absolute distance between the predicted and ground truth landmarks, 
\(x^{\text{pred}}\) and \(x^{\text{gt}}\) are the predicted and ground truth \(x\)-coordinates, respectively, 
and similarly for \(y^{\text{pred}}\) and \(y^{\text{gt}}\), \(N\) denotes the total number of landmarks.

\textbf{Success Detection Rate (SDR):}
The proportion of landmarks accurately detected within specified radius thresholds is used to evaluate detection accuracy. For the head datasets, SDR is calculated at 2.0mm, 2.5mm, 3.0mm, and 4.0mm. For the hand dataset, SDR is assessed at 2.0mm, 4.0mm, and 10.0mm, while for the pelvic dataset, it is calculated at 2px, 2.5px, 3.0px, and 4.0px.
\begin{equation}
\hspace{1cm} \text{SDR}_z = \frac{\# \{ j : R_i < z \}}{N} \times 100\%,
\end{equation}
where \(\text{SDR}_z\) represents the SDR within a threshold \(z\),
the term \(\# \{ j : R_i < z \}\) counts the number of landmarks within the threshold \(z\).

\subsection{Baseline Methods}
To verify the effectiveness of our proposed model, we compare it with several representative or SOTA methods, including U-Net~\cite{Unet}, SCN~\cite{SCN}, Cascade R-CNN~\cite{CascadeRCNN}, GU2Net~\cite{GU2Net}, DATR~\cite{DATR}, CeLDA~\cite{CELDA}, SR-UNet~\cite{SRLD-Net}, FARNet~\cite{FARNet}, HTC~\cite{HTC}:

\begin{itemize}
    \item \textbf{U-Net}\cite{Unet}: A model for medical image segmentation with an encoder-decoder structure and skip connections, enhancing accuracy and robustness.
    \item \textbf{SCN}\cite{SCN}: Combines locally accurate candidate heatmaps with globally consistent spatial configuration heatmaps, improving detection under limited data.
    \item \textbf{Cascade R-CNN}\cite{CascadeRCNN}: A multi-stage framework that enhances object detection by progressively increasing IoU thresholds.
    \item \textbf{GU2Net}\cite{GU2Net}: Integrates local and global networks with dilated convolutions, improving landmark detection accuracy across diverse anatomical regions.
    \item \textbf{DATR}\cite{DATR}: Uses transformers for landmark detection across multiple anatomical regions, with domain-adaptive transformers to enhance accuracy in multi-domain scenarios.
    \item \textbf{CeLDA}\cite{CELDA}: Uses prototype networks to address visual differences across age groups and leverages anatomical relationships for precise landmark detection.
    \item \textbf{SR-UNet}\cite{SRLD-Net}: Applies super-resolution networks and pyramid pooling to reduce errors and improve detection precision.
    \item \textbf{FARNet}\cite{FARNet}: Utilizes multi-scale feature aggregation and refinement to generate high-resolution heatmaps, significantly boosting detection accuracy.
    \item \textbf{HTC}\cite{HTC}: Combines multi-resolution learning and a hybrid Transformer-CNN architecture to improve landmark detection accuracy.

\end{itemize}

\subsection{Implementation Details}
In experiments, we utilize the MMPose framework from OpenMMLab, an open-source pose estimation toolkit based on PyTorch that provides extensive model and dataset interfaces suitable for landmark detection tasks. The model is implemented using PyTorch 1.9.0 and Python 3.9 and runs on 4 RTX 3090 GPUs in a CUDA 11 environment. 

The inputs are subjected to data augmentation, and multi-heatmap deep supervision learning is employed. The standard deviations for the three levels of deep supervision heatmaps are set at \( \sigma_1 = 2 \), \( \sigma_2 = 2 \), and \( \sigma_3 = 4 \). The MSE loss function is used, with deep supervision loss weights set to \( w_1 = 1 \), \( w_2 = 3 \), and \( w_3 = 3 \). The model is trained using the AdamW optimizer with a learning rate of \( 4 \times 10^{-4} \) and a weight decay of 0.01 for 150 epochs. 


\subsection{Experiment Results}
To validate the broad applicability of the proposed method across different anatomical regions and datasets, we organize the experimental results into groups based on head, hand, and pelvic regions, ensuring clarity and ease of understanding. Specifically, Section~\ref{sec:res_head} presents the results for the head datasets (ISBI2015, ISBI2023, and the CephAdoAdu dataset), Section~\ref{sec:res_hand} reports the results for the hand dataset, and Section~\ref{sec:res_pelvic} details the results for the pelvic dataset. Each set of results is compared against U-Net and current SOTA methods (HTC and FARNet) to comprehensively assess the advantages of our approach.

\subsubsection {Head dataset results}
\label{sec:res_head}

For the head datasets, we conduct experiments using the two most widely used datasets, ISBI2015 and ISBI2023, as well as a private dataset, CephAdoAdu. As shown in Table \ref{tab:head_results}, the results indicate that our method achieves optimal MRE on both the ISBI2015 and ISBI2023 datasets, with MREs of 1.13mm and 1.05mm, and STDs of 1.11 and 1.73, respectively. The improvement in performance in MRE is particularly pronounced on the ISBI2015 dataset, which has a smaller training sample size, demonstrating a 0.05mm decrease compared to the previously SOTA method HTC (approximately 5\% reduction), which indicates the significant effectiveness of our method in data-limited scenarios, showcasing its potential in low-data environments.

\begin{figure*}[htbp]
\centering
\begin{minipage}{\textwidth} 
  \centering
  \includegraphics[width=\linewidth,trim={0cm 0cm 0cm 0cm},clip]{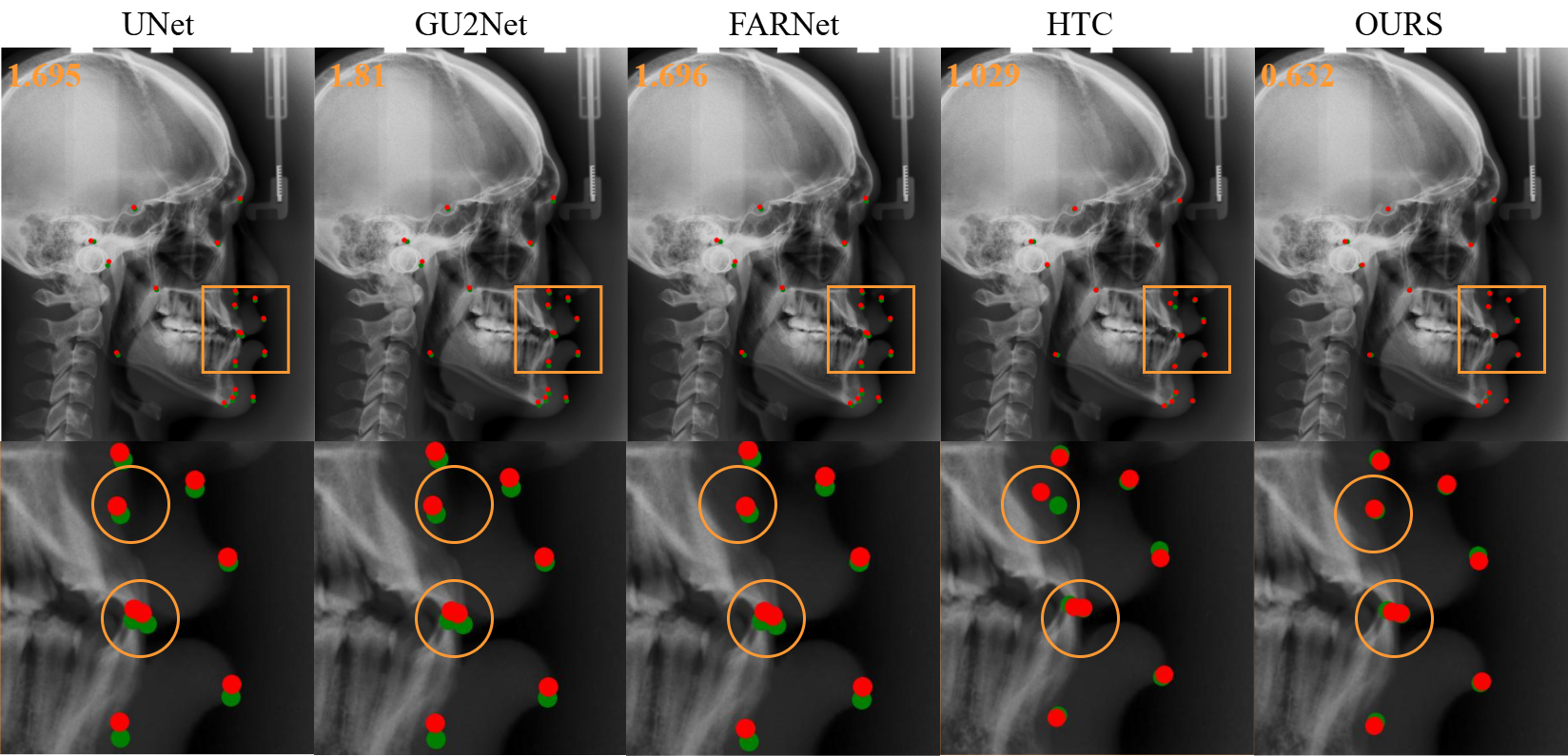}
  \captionsetup{width=\textwidth} 
  \caption{Visualizations of various methods on the ISBI2015 dataset. The red points represent the predicted landmarks, while the green points correspond to the ground truth labels. Local details are provided below for a clearer comparison of the results. The MRE value is shown in the top-left corner for reference.}
  \label{fig: ISBI2015}
\end{minipage}
\end{figure*}

On the ISBI2023 dataset, our method further reduces the MRE by 0.03mm. The ISBI2023 dataset includes images from various devices and annotations for 29 landmarks, demonstrating that our approach is well-suited for datasets with a higher number of landmarks and variability in image sources. At different SDR thresholds (e.g., SDR 2.0mm and SDR 2.5mm), our method outperforms SOTA models like HTC and FARNet on the ISBI2023 dataset. Specifically, the SDR at 2.0mm and 2.5mm reaches 84.78\% and 89.92\%, surpassing HTC by 1.54\% and 0.72\%, and outperforming FARNet by 2.27\% and 1.35\%, respectively.

\begin{figure*}[htbp]
\centering
\begin{minipage}{\textwidth} 
  \centering
  \includegraphics[width=\linewidth, trim={0cm 0cm 0cm 0cm}, clip]{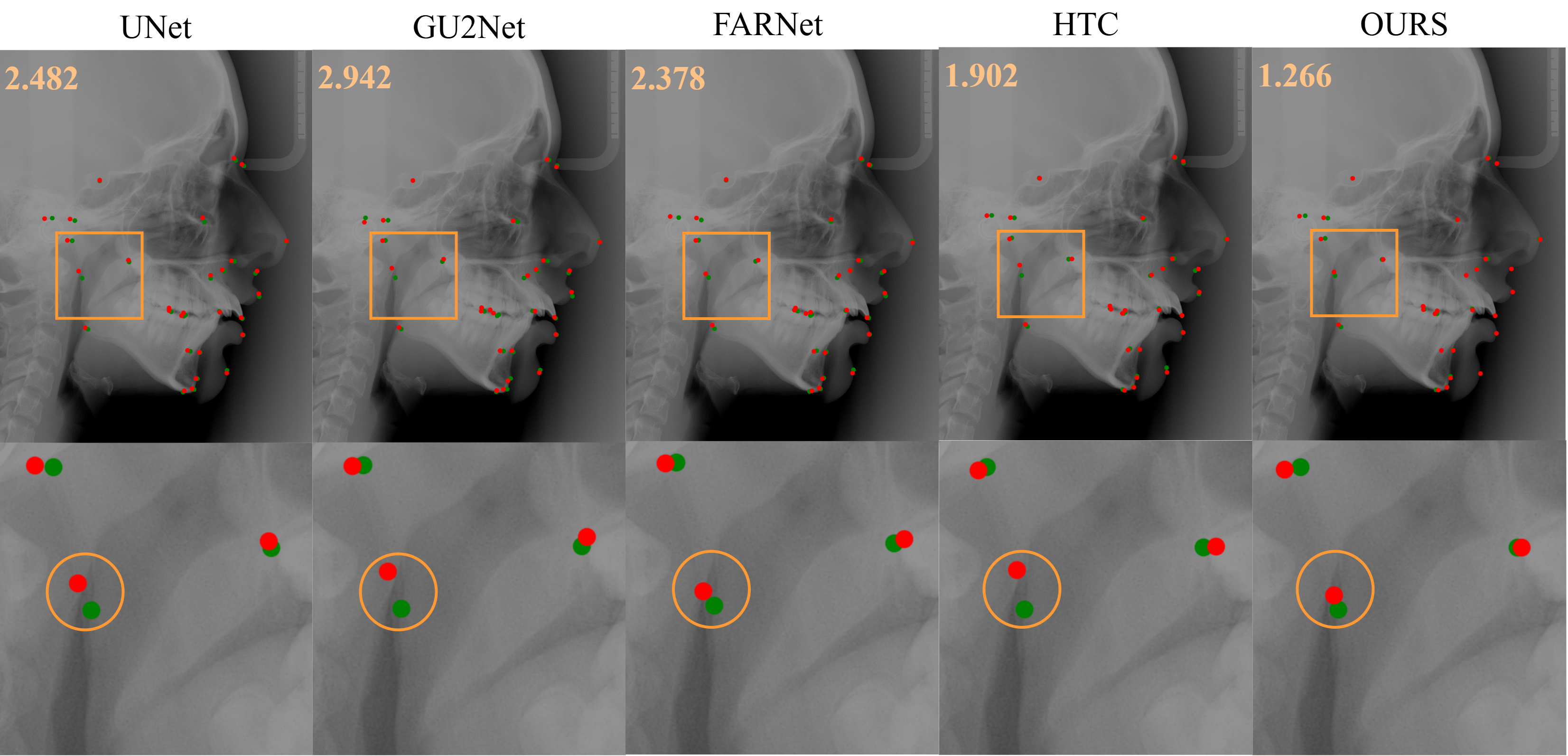}
\end{minipage}
\caption{Visualizations of various methods on the ISBI2023 dataset.}
\label{fig: ISBI2023}
\end{figure*}

The visualizations of different landmark detection methods on the ISBI2015 and ISBI2023 datasets are shown in Fig.\ref{fig: ISBI2015} and Fig.\ref{fig: ISBI2023}. The four methods compared are U-Net, GU2Net, FARNet, and HTC. It is clear that the overall MRE has decreased. Notably, HTC and FARNet show greater localization errors in the mandibular and cervical regions, whereas our method demonstrates a marked improvement in the alignment of predicted and actual landmarks in these areas. The orange rectangles and circles highlight these differences, showcasing that our method achieves higher accuracy and consistency in landmark detection within complex skeletal structures.

As shown in Table \ref{tab:head_results}, the results on the CephAdoAdu dataset also demonstrate outstanding performance, with an average MRE of 0.98mm. This represents a 0.07mm reduction (over 7\%) compared to the previous SOTA method, CeLDA, which was specifically designed for this dataset and outperforms earlier general models such as GU2Net and other SOTA methods like HTC. Although our method is not specially designed for this dataset, it still achieves significant improvements, highlighting its broad applicability.

\begin{table*}[htbp]
\centering
\caption{Comparison of MRE and SDR results on ISBI2015, ISBI2023, and CELDADataset. The best results are in bold, $*$ indicates the result is taken from referenced papers.}
\begin{tabular}{llccccc}
\hline
\multirow{2}{*}{\textbf{Dataset}} & \multirow{2}{*}{\textbf{Model}} & \multirow{1}{*}{\textbf{MRE} $\pm$ \textbf{STD}}& \multicolumn{4}{c}{\textbf{SDR (\%)}} \\ 
  &  & \textbf{(mm)} & \textbf{2mm} & \textbf{2.5mm} & \textbf{3mm} & \textbf{4mm} \\ 
\hline
\multirow{6}{*}{\textbf{ISBI2015}}
& U-Net~\cite{Unet} & $ 2.05\pm 2.69$ & 59.28 & 77.24 & 86.80 & 95.13 \\
& CELDA~\cite{CELDA} & $1.37 \pm 0.41$ & 79.57 & 87.74 & 91.95 & 96.67 \\
& GU2Net~\cite{GU2Net}* & $1.54 \pm 0.51$ & 77.79 & 84.65 & 89.41 & 94.93 \\
& FARNet~\cite{FARNet}* & $1.27 \pm 0.47$ & 82.51 & 88.58 & 92.71 & 96.84 \\
& HTC~\cite{HTC}  & $1.19 \pm 1.37$ & 83.24 & 89.20 & 92.88 & \textbf{97.01} \\\gr
& \textbf{Ours} & $\textbf{1.13} \pm \textbf{1.11}$ & \textbf{84.78} & \textbf{89.92} & \textbf{92.99} & 96.76 \\
\hline
\multirow{6}{*}{\textbf{ISBI2023}}
& U-Net~\cite{Unet} & $1.58 \pm 5.01$ & 78.62 & 84.81 & 88.39 & 93.52 \\
& CELDA~\cite{CELDA}  & $1.64 \pm 0.63$ & 78.00 & 85.00 & 90.60 & 95.38 \\
& GU2Net~\cite{GU2Net}* & $1.78 \pm 5.10$ & 81.97 & 86.31 & 88.87 & 92.28 \\
& FARNet~\cite{FARNet} & $1.10 \pm 1.33$ & \textbf{87.59} & \textbf{91.88} & \textbf{94.48} & \textbf{96.87} \\
& HTC~\cite{HTC} & $1.08 \pm 1.81$ & 87.11 & 91.11 & 93.70 & 96.69 \\\gr
& \textbf{Ours} & $\textbf{1.05} \pm \textbf{1.73}$ & 87.43 & 91.43 & 93.70 & 96.59 \\
\hline
\multirow{8}{*}{\textbf{CELDA (Average)}}
& U-Net~\cite{Unet} & $1.56 \pm 5.39$ & 81.30 & 88.20 & 92.93 & 96.66 \\
& Cascade RCNN~\cite{CascadeRCNN}* & $2.31 \pm 0.94$ & 61.47 & 73.20 & 81.13 & 90.77 \\
& SCN~\cite{SCN}* & $1.59 \pm 1.73$ & 82.97 & 90.34 & 93.47 & 95.37 \\
& GU2Net~\cite{GU2Net}* & $1.69 \pm 0.91$ & 80.33 & 88.13 & 91.94 & 95.57 \\
& SR-UNet~\cite{SRLD-Net}* & $1.40 \pm 0.93$ & 87.17 & 91.81 & 94.31 & 96.70 \\
& HTC~\cite{HTC} & $1.11 \pm 1.08$ & 88.36 & 91.94 & 94.43 & 97.10 \\
& CeLDA~\cite{CELDA} * & $1.05 \pm 0.33$ & \textbf{89.13} & \textbf{93.60} & \textbf{99.67} & \textbf{98.67} \\\gr
& \textbf{Ours} & $\textbf{0.98} \pm \textbf{0.33}$ & 88.43 & 93.00 & 95.13 & 98.00 \\
\hline
\multirow{8}{*}{\textbf{CELDA (Adult)}}
& U-Net~\cite{Unet} & $1.56 \pm 3.82$ & 77.00 & 85.60 & 91.93 & 96.46 \\
& Cascade RCNN~\cite{CascadeRCNN}* & $2.19 \pm 0.97$ & 59.93 & 72.13 & 80.47 & 90.80 \\
& SCN~\cite{SCN}* & $1.53 \pm 1.75$ & 82.37 & 89.86 & 93.00 & 95.73 \\
& GU2Net~\cite{GU2Net}* & $1.46 \pm 0.93$ & 82.07 & 88.80 & 92.07 & 96.33 \\
& SR-UNet~\cite{SRLD-Net}* & $1.13 \pm 0.89$ & 86.18 & 91.25 & 94.03 & 97.33 \\
& HTC~\cite{HTC} & $1.11 \pm 1.09$ & 85.60 & 91.80 & 94.31 & 97.43 \\
& CeLDA~\cite{CELDA} * & $1.10 \pm 1.17$ & \textbf{88.33} & \textbf{93.00} & \textbf{95.67} & 97.53 \\\gr
&\textbf{Ours} & $\textbf{1.00} \pm \textbf{0.70}$ & 85.53 & 91.20 & 94.00 & \textbf{97.53} \\
\hline
\multirow{8}{*}{\textbf{CELDA (Teenager)}}
& U-Net~\cite{Unet} & $1.56 \pm 6.60$ & 85.60 & 90.80 & 93.93 & 96.86 \\
& Cascade RCNN~\cite{CascadeRCNN}* & $2.43 \pm 0.94$ & 60.37 & 73.87 & 81.18 & 90.73 \\
& SCN~\cite{SCN}* & $1.50 \pm 1.70$ & 83.87 & 90.97 & 94.17 & 95.98 \\
& GU2Net~\cite{GU2Net}* & $1.55 \pm 1.10$ & 80.33 & 88.13 & 91.94 & 95.57 \\
& SR-UNet~\cite{SRLD-Net}* & $1.17 \pm 0.99$ & 87.73 & 94.33 & 96.80 & 98.33 \\
& HTC~\cite{HTC} & $1.03 \pm 1.07$ & 91.10 & 94.31 & 96.33 & 98.73 \\
& CeLDA~\cite{CELDA} * & $1.05 \pm 0.90$ & 89.93 & 94.27 & \textbf{96.93} & \textbf{98.63} \\\gr
&\textbf{Ours} & $\textbf{0.85} \pm \textbf{0.95}$ & \textbf{91.33} & \textbf{95.00} & 96.90 & 98.33 \\
\hline
\end{tabular}
\label{tab:head_results}
\end{table*}

\begin{figure*}[htbp]
\centering
\begin{minipage}{\textwidth}
  \centering
  \includegraphics[width=\linewidth,trim={0cm 0cm 0cm 0cm},clip]{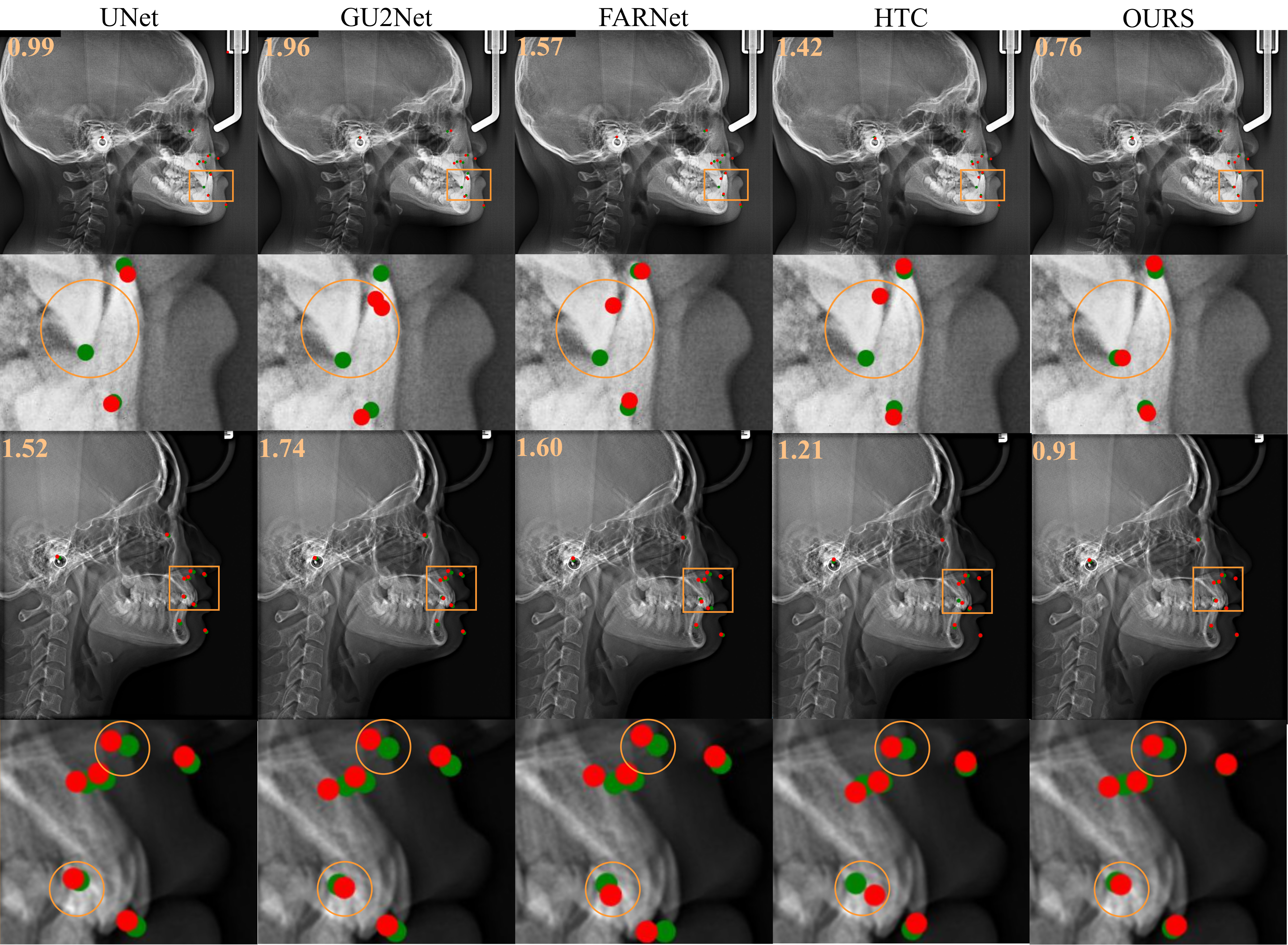}
\end{minipage}
\caption{Visualizations of various methods on the CephAdoAdu dataset.}
\label{fig: CELDA}
\end{figure*}

Following the approach in the CeLDA~\cite{CELDA}, we further report results separately for adolescent and adult data. Our method demonstrates exceptional performance on the adolescent subset, achieving an MRE of 0.85px, which is 15\% reduction compared to the SOTA result. Additionally, SDR improves across all thresholds, with the SDR within 2mm increasing from 89.93\% to 91.33\%. When compared to other general methods such as U-Net, GU2Net, and SR-UNet, our method consistently shows superior SDR at all detection thresholds, with more pronounced improvements. As illustrated in Fig.~\ref{fig: CELDA}, there are clear anatomical differences between the left-side adolescent image and the right-side adult image, yet our method accurately predicts landmarks in both cases.

\subsubsection{Hand Dataset Results}\label{sec:res_hand}
In the experiments on the hand dataset, we conduct a comprehensive comparison with HTC, which has SOTA results to date, as well as other methods such as CELDA, DATR, GU2Net, and FARNet. The results in Table~\ref{tab:hand_results} demonstrate that our method achieves significant superiority across all evaluation metrics. Our MRE is only 0.53mm, representing a reduction of 0.033mm compared to HTC and 0.14mm compared to the FARNet network. The SDR at thresholds of \(\leq\)2.0mm and \(\leq\)4.0mm reach 97.09\% and 99.7\%, respectively, showing improvements of 0.25\% and 0.07\% over HTC.

Considering that previous results on this dataset are already highly accurate, the fact that our method still achieves improvements highlights its significant advantage in detecting fine anatomical structures in high-resolution images, particularly for hand landmark detection. This further validates the generalizability and robustness of our method.

\begin{table}[htbp]
\centering
\caption{Comparison of MRE and SDR results on the Hand dataset. $*$ indicates the result is taken from referenced papers. $-$ means the result is missing in the referenced paper.}
\resizebox{\columnwidth}{!}{%
\begin{tabular}{lcccc}
\hline
 \multirow{2}{*}{\textbf{Model}} & \multirow{1}{*}{\textbf{MRE} $\pm$ \textbf{STD}}& \multicolumn{3}{c}{\textbf{SDR (\%)}} \\ 
   & \textbf{(mm)} & \textbf{2mm}  & \textbf{4mm} & \textbf{10mm} \\
\hline
U-Net~\cite{Unet} & 0.89 \(\pm\) 0.89 & 95.69 & 99.26 & 99.65 \\
DATR~\cite{DATR}* & 0.86 \(\pm\) - & 94.04 & 99.20 & 99.31 \\
GU2Net~\cite{GU2Net}* & 0.63 \(\pm\) 1.36 & 96.01 & 99.39 & 99.98 \\
SCN~\cite{SCN}* & 0.66 \(\pm\) 0.74 & 94.99 & 99.27 & 99.99 \\
FARNet~\cite{FARNet} & 0.67 \(\pm\) 0.74 & 95.65 & 99.58 & 99.99 \\
CeLDA~\cite{CELDA} & 0.70 \(\pm\) 0.77 & 95.26 & 99.40 & 99.99 \\
HTC~\cite{HTC}* & 0.56 \(\pm\) 0.58 & 96.84 & 99.63 & 100.00 \\
\gr\textbf{Ours} & \textbf{0.53} \(\pm\) \textbf{0.56} & \textbf{97.09} & \textbf{99.70} & \textbf{100.00} \\
\hline
\end{tabular}%
}
\label{tab:hand_results}
\end{table}

\begin{figure*}[htbp]
\centering
\begin{minipage}{\textwidth}
  \centering
  \includegraphics[width=\linewidth,trim={0cm 0cm 0cm 0cm},clip]{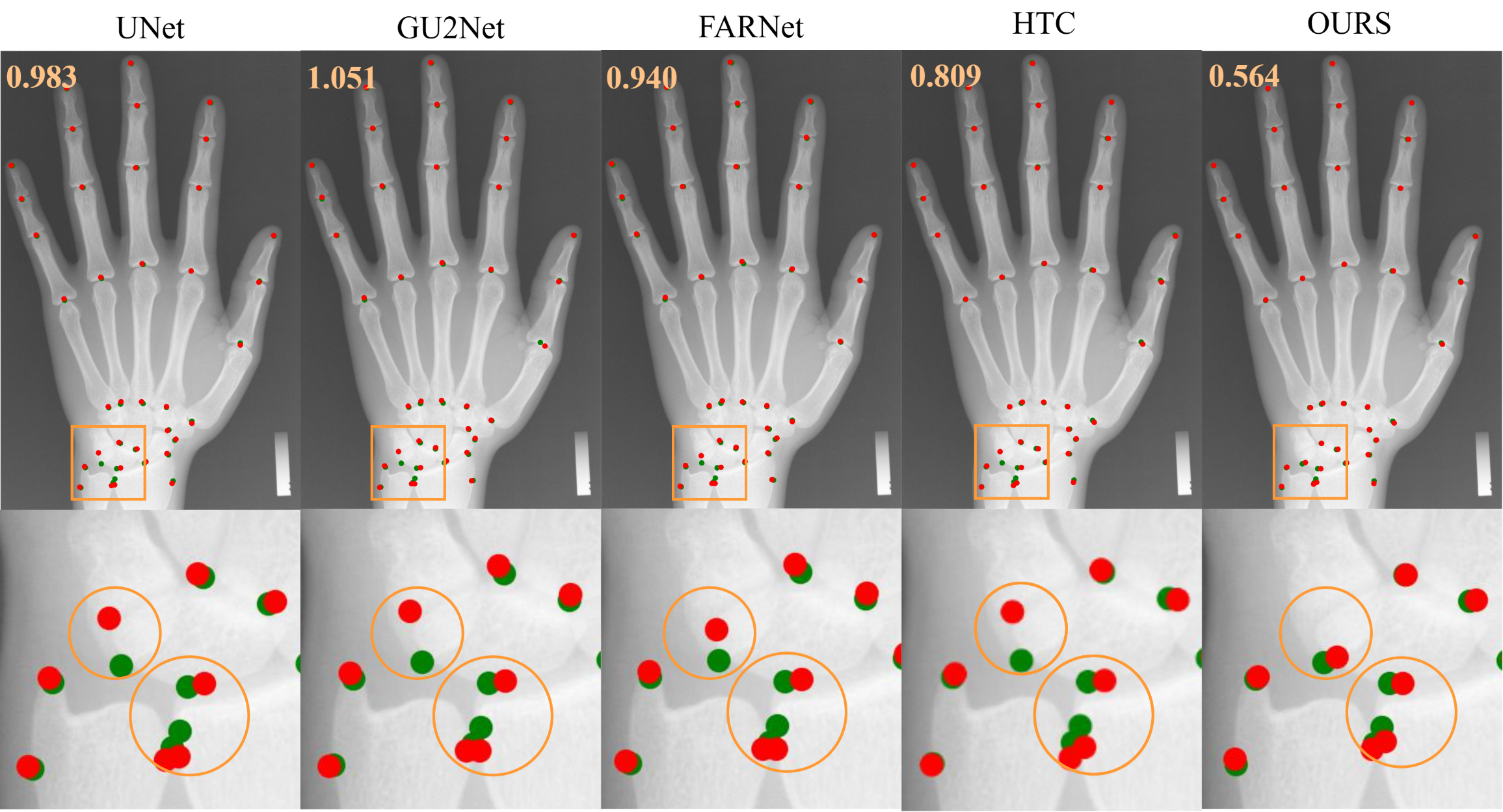}
\end{minipage}
\caption{Visualizations of various methods on the Hand X-Rays dataset.}
\label{fig: hand}
\end{figure*}

Fig.~\ref{fig: hand} shows the visual results on the hand dataset. The comparison highlights that methods such as HTC exhibit notable localization errors, especially in the wrist and joint regions. In contrast, our model significantly enhances the accuracy of landmark detection. The zoomed-in view, enclosed within the orange rectangle, emphasizes the areas where our model excels, showcasing higher precision and consistency. This is particularly evident in the detection of finer hand joint structures, where our model achieves more accurate results.

\subsubsection{Pelvic Dataset Results}\label{sec:res_pelvic}
In the experiments on the pelvic dataset, our method also achieves SOTA performance, demonstrating significant improvements over previous SOTA methods. As shown in Table~\ref{tab:pelvic_results}, our method's MRE is only 6.64px, representing a reduction of 2.63px compared to FARNet and 1.09px compared to HTC. The highest SDR within a 2px threshold reaches 26\%, surpassing HTC by 7.2\% and FARNet by 0.8\%. Overall, our model achieves both a lower MRE and a higher SDR.

In comparison to general methods like U-Net and GU2Net, our method demonstrates particularly remarkable improvements, with the MRE being approximately one-third of those methods and SDR showing significant gains. According to PELE~\cite{PELE}, landmark recognition in the pelvic region is especially challenging due to complex structures and increased occlusions. Nonetheless, our method achieves SOTA results with notable enhancements, underscoring its broad applicability and potential for real-world applications.

\begin{table}[htbp]
\centering
\caption{Comparison of MRE and SDR results on the Pelvic dataset.}
\resizebox{\columnwidth}{!}{%
\begin{tabular}{lccccc}
\hline
 \multirow{2}{*}{\textbf{Model}} & \multirow{1}{*}{\textbf{MRE} $\pm$ \textbf{STD}}& \multicolumn{4}{c}{\textbf{SDR (\%)}} \\ 
   & \textbf{(px)} & \textbf{2px} & \textbf{2.5px} & \textbf{3px} & \textbf{4px} \\ 
\hline
U-Net~\cite{Unet} & 18.58 \(\pm\) 42.77 & 23.60 & 34.80 & 42.80 & 54.40  \\
GU2Net~\cite{GU2Net} & 22.28 \(\pm\) 52.77 & 22.00 & 32.80 & 41.20 & 51.20  \\
FARNet~\cite{FARNet} & 9.27 \(\pm\) 15.60 & 25.20 & \textbf{35.60} & 39.60 & 56.60  \\
HTC~\cite{HTC} & 7.23 \(\pm\) 16.12 & 18.80 & 28.80 & 37.20 & 54.80  \\
\gr\textbf{Ours} & \textbf{6.64 \(\pm\) \textbf{14.82}} & \textbf{26.00} & 35.20 & \textbf{45.60} & \textbf{60.40} \\
\hline
\end{tabular}%
}
\label{tab:pelvic_results}
\end{table}

\begin{figure*}[htbp]
\centering
\begin{minipage}{\textwidth}
  \centering
  \includegraphics[width=\linewidth,trim={0cm 0cm 0cm 0cm},clip]{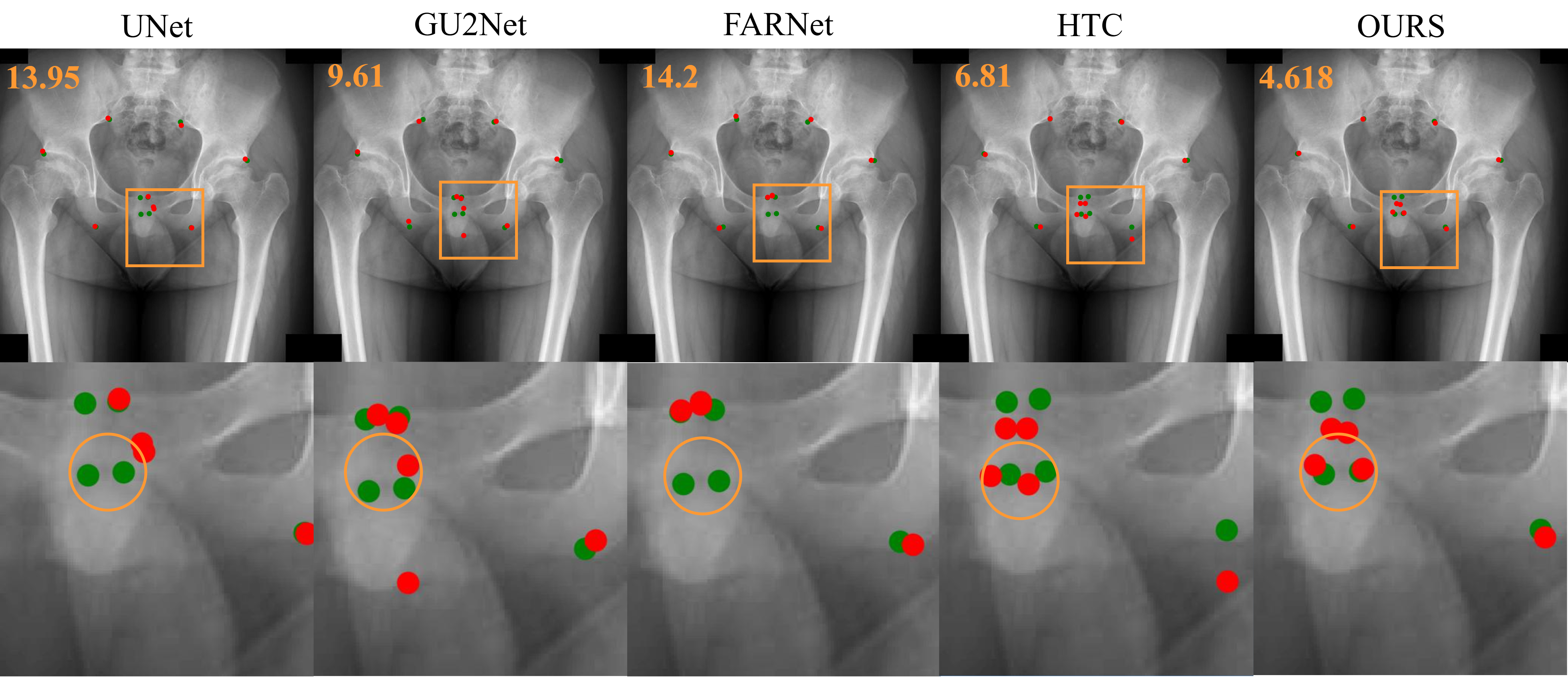}
\end{minipage}
\caption{Visualizations of various methods on the Pelvic X-Ray dataset.}
\label{fig:pelvic}
\end{figure*}


Overall, our method achieves significant performance improvements across multiple datasets, surpassing existing methods and fully reaching SOTA. Grouped reporting of results for the head, hand, and pelvic datasets provides a clear visualization of the experimental outcomes for each anatomical region, confirming the method's applicability and robustness. Additionally, the extensive experiments we conduct demonstrate that our method performs well under varying resolutions, image qualities, sampling conditions, different dataset characteristics, and noise levels, showcasing excellent practicality and stability.

\subsection{Ablation Study}
On the ISBI2015 dataset, we evaluate the impact of incorporating the CBAM into the CNN module and adding an FFCM on the model's performance. Initially, we integrate the CBAM module into the CNN architecture to enhance its feature extraction capabilities. As shown in Table~\ref{tab:ablation}, the addition of CBAM reduces the MRE by 0.007mm and leads to slight improvements in the SDR, indicating that CBAM can improve the precision of landmark detection.

\begin{table}[htbp]
\centering
\caption{Quantitative Performance Analysis of Ablation Study on CBAM and FFCM: Comparing MRE and SDR.}
\resizebox{\columnwidth}{!}{%
\begin{tabular}{lccccc}
\hline
 \multirow{2}{*}{\textbf{Model}} & \multirow{1}{*}{\textbf{MRE} $\pm$ \textbf{STD}}& \multicolumn{4}{c}{\textbf{SDR (\%)}} \\ 
   & \textbf{(mm)} & \textbf{2mm} & \textbf{2.5mm} & \textbf{3mm} & \textbf{4mm} \\ 
\hline
BiFormer & 1.147 \(\pm\) 1.17 & 84.27 & 89.45 & 93.07 & 97.01  \\
BiFormer+CBAM & 1.140 \(\pm\) 1.18 & 84.28 & 89.64 & \textbf{93.12} & \textbf{97.02}  \\
\gr BiFormer+CBAM+FFCM & \textbf{1.135} \(\pm\) \textbf{1.11} & \textbf{84.78} & \textbf{89.92} & 92.99 & 96.76  \\
\hline
\end{tabular}%
}
\label{tab:ablation}
\end{table}

Following this, we further add an FFCM to the CBAM-enhanced model to improve the capture of global information. As shown in Table~\ref{tab:ablation}, this modification leads to additional improvements in both the MRE and SDR metrics, with the MRE decreasing by an additional 0.005mm. 

\begin{table*}[htbp]
\centering
\caption{Quantitative Performance Analysis of Different Transformer Architectures as Backbone: Comparing MRE and SDR on ISBI2015, ISBI2023, and Hand X-ray Datasets. Swin-T refers to Swin Transformer, and Basic refers to the Transformer architecture.}
\begin{tabular}{llccccc}
\hline
 \multirow{2}{*}{\textbf{Dataset}} & \multirow{2}{*}{\textbf{Backbone}} & \multirow{1}{*}{\textbf{MRE} $\pm$ \textbf{STD}}& \multicolumn{4}{c}{\textbf{SDR (\%)}} \\ 
   & &\textbf{(mm)} & \textbf{2mm} & \textbf{2.5mm} & \textbf{3mm} & \textbf{4mm} \\ 
\hline
\multirow{3}{*}{\textbf{ISBI2015}}
& Basic & $1.186 \pm 1.37$ & 83.24 & 89.20 & 92.88 & 97.01 \\
& Swin-T & $1.162 \pm 1.09$ & 84.22 & 89.42 & 93.06 & 97.18 \\\gr
& \textbf{BiFormer} & $\textbf{1.146} \pm \textbf{1.07}$ & \textbf{84.27} & \textbf{89.45} & \textbf{93.07} & \textbf{97.41} \\
\hline
\multirow{3}{*}{\textbf{ISBI2023}}
& Basic & $1.083 \pm 1.80$ & 86.70 & 90.90 & 93.48 & 96.49 \\
& Swin-T & $1.068 \pm 1.81$ & 86.76 & 90.76 & \textbf{93.66} & 96.51\\\gr
& \textbf{BiFormer} & $\textbf{1.055} \pm \textbf{1.74}$ & \textbf{87.39} & \textbf{90.98} & 93.48 & \textbf{96.59} \\
\hline
\multirow{3}{*}{\textbf{Hand}}
& Basic & $0.575 \pm 0.55$ & 97.37 & 98.71 & 99.30 & 99.62 \\
& Swin-T & $0.569 \pm 0.57$ & 97.42 & \textbf{98.75} & \textbf{99.38} & 99.63 \\\gr
& \textbf{BiFormer} & \textbf{0.539} \(\pm\) \textbf{0.55} & \textbf{97.43} & 98.73 & 99.31 & \textbf{99.66} \\
\hline
\end{tabular}
\label{tab:ablation2}
\end{table*}

Table~\ref{tab:ablation2} presents a comprehensive comparison of different Transformer architectures as backbone across three medical imaging datasets (ISBI2015, ISBI2023, and Hand X-rays), evaluating their performance through MRE and SDR. The BiFormer architecture demonstrates superior performance with the lowest MRE values across all datasets: 1.146 ± 1.07mm for ISBI2015, 1.055 ± 1.74mm for ISBI2023, and 0.539 ± 0.55mm for Hand dataset. In terms of SDR, which measures detection accuracy at various threshold distances (2mm, 2.5mm, 3mm, and 4mm), BiFormer consistently achieves higher success rates. For instance, on ISBI2023, BiFormer achieves an SDR of 87.39\% at 2mm threshold, surpassing both basic Transformer (86.70\%) and Swin Transformer (86.76\%).

The experimental results validate the effectiveness of our architectural improvements. Using BiFormer as the backbone consistently outperforms Transformer and Swin Transformer models across multiple metrics and datasets, with SDR improving by over 1\% and MRE decreasing by more than 4\%. Adding the CBAM and FFCM modules further enhances SDR and reduces MRE. These improvements confirm that BiFormer provides a strong foundation, CBAM strengthens feature attention, and FFCM improves global information capture. The consistent performance across datasets highlights the robustness and generalization ability of our method, making it more accurate and reliable for medical landmark detection compared to existing methods.


\section{Conclusion}
This paper presents the Hybrid Attention Network(HYATT-Net) for accurate and efficient ALD in medical images. This novel architecture combines CNNs and Transformers, using a dynamic sparse attention mechanism (BiFormer with Bi-Level Routing Attention) to efficiently handle high-resolution images. The BiFormer captures global context, while Attention Residual Blocks (ARMs), enhanced by CBAM, refine local features. A Feature Fusion Correction Module (FFCM) integrates multi-scale features, preventing resolution loss. Extensive experiments across five diverse datasets demonstrate state-of-the-art performance, exceeding existing methods in accuracy, robustness, and efficiency. The HYATT-Net offers a promising framework for various medical image analysis tasks. Future work will explore 3D image applications and further optimize the sparse attention mechanism. This method has the potential to significantly improve clinical practice through more accurate and efficient image-guided procedures.

\bibliographystyle{unsrt}

\bibliography{cas-dc}




\end{document}